%% file: main.tex
\pdfoutput=1

\documentclass[11pt]{article}

\usepackage[]{EMNLP2022}

\usepackage{times}
\usepackage{latexsym}

\usepackage[T1]{fontenc}

\usepackage[utf8]{inputenc}

\usepackage{microtype}

\usepackage{inconsolata}

\usepackage{times}
\usepackage{latexsym}
\usepackage{amsmath}
\usepackage{amsthm}
\usepackage{booktabs}
\usepackage{multirow}
\usepackage{graphicx}
\usepackage{threeparttable}

\usepackage{enumitem}
\usepackage{float}
\usepackage{amsfonts}
\usepackage[ruled,vlined]{algorithm2e}
\usepackage{algorithmic}

\usepackage[T1]{fontenc}
\usepackage{array}
\newcolumntype{C}[1]{>{\centering\arraybackslash}p{#1}}
\newcolumntype{P}{>{\centering\arraybackslash}p{2.5em}}

\title{Coordinated Topic Modeling}

\author{Pritom Saha Akash $\quad$ Jie Huang $\quad$ Kevin Chen-Chuan Chang \\
University of Illinois at Urbana-Champaign, USA \\
 \texttt{\{pakash2, jeffhj, kcchang\}@illinois.edu}
}

\begin{document}
\maketitle
\input{EMNLP 2022/sections/abstract}

\input{EMNLP 2022/sections/introduction}

\input{EMNLP 2022/sections/methodology}

\input{EMNLP 2022/sections/experiments}

\input{EMNLP 2022/sections/conclusion}

\input{EMNLP 2022/sections/acknowledgement}

\input{EMNLP 2022/sections/limitations}

\bibliography{anthology}
\bibliographystyle{acl_natbib}

\input{EMNLP 2022/sections/appendix}

\end{document}

%% file: EMNLP 2022/sections/abstract.tex
\begin{abstract}

We propose a new problem called coordinated topic modeling that imitates human behavior while describing a text corpus. It considers a set of well-defined topics like the axes of a semantic space with a reference representation. It then uses the axes to model a corpus for easily understandable representation. This new task helps represent a corpus more interpretably by reusing existing knowledge and benefits the corpora comparison task. We design ECTM, an embedding-based coordinated topic model that effectively uses the reference representation to capture the target corpus-specific aspects while maintaining each topic's global semantics. In ECTM, we introduce the topic- and document-level supervision with a self-training mechanism to solve the problem. Finally, extensive experiments on multiple domains show the superiority of our model over other baselines.\footnote{Code and data are available at \href{https://github.com/pritomsaha/Coordinated-Topic-Modeling}{https://github.com/ pritomsaha/Coordinated-Topic-Modeling}}

\end{abstract}

%% file: EMNLP 2022/sections/introduction.tex
\section{Introduction}
\label{sec:intro}
We often ask questions about well-defined topics when we read articles (e.g., news/research articles). E.g., in news domain, such a question can be like ``what is in \textit{entertainment} today? (it's about \textit{oscar 2022})'' or in academic domain, it can be like ``what is trending in \textit{machine learning} in 2016? (it's about \textit{deep learning})''. This is a practical human trait while understanding information in a text corpus. Rather than finding arbitrary topics, people often want to explore the text based on some well-defined topics. E.g., in news domain, such topics are \textit{business}, \textit{politics}, \textit{sports}, etc.

A well-defined topic is not merely a name (i.e., surface name); it generally also has a representation (e.g., word distribution) that can be obtained from a large text corpus, which we may call the \textit{reference representation}. E.g., in such representation of the topic \textit{sports}, words like \textit{play}, \textit{game}, and \textit{player} are mainly dominated. Similarly, for \textit{politics}, these are \textit{vote}, \textit{election}, \textit{party}, etc. 
However, a topic's representation generally deviates from its reference to other corpora. E.g., while describing a news corpus specific to the USA, it is likely to use words such as \textit{nfl}, \textit{baseball} or \textit{football} to represent \textit{sports}, while words like \textit{cricket}, \textit{ipl} or \textit{run} are more relevant for India.

\input{EMNLP 2022/figures/example}

The above scenario can be explained by the concept of the coordinate system in geometry. Specifically, we can consider a set of well-defined topics as the axes or basis of semantic space with their reference representation $\beta^{r}$. E.g., in Figure \ref{fig:example}, two topics $\mathcal{C}_1$:\textit{sports} and $\mathcal{C}_2$:\textit{politics} represents two axes (i.e., reference axes) defined by their  reference representations $\beta^r_1$ and $\beta^r_2$ respectively. Defining a target corpus by these two topics is analogous to finding a representation in the space defined by the reference axes. Figure \ref{fig:example} shows such target representation $\beta^{USA}$ and $\beta^{INDIA}$ as coordinates by reference axes for corpora specific to USA and INDIA, respectively.

In practice, the above interpretation of defining a corpus can be helpful from several perspectives. 
First, it has better \textbf{interpretability} than traditional document modeling approaches like topic models \cite{blei2003latent,jordan1999introduction} that discover some unknown topics. It defines a corpus over well-defined topics by finding their corpus-specific representations. Thus, users do not need much effort to understand unknown topics; instead, they easily grasp the corpus-specific aspects of given well-known topics. Second, it facilitates the \textbf{comparability} between two corpora. It uses a predefined set of topics with their reference representation as the axes to describe a corpus in a semantic space. Thus, two corpora represented by the same axes can be easily comparable by studying their relative position in the coordinates. Third, it allows \textbf{reusability} by utilizing existing information to explore new knowledge. As the reference representation is already known information about some topics, resuing it to model a new corpus in a similar domain is intuitive and helpful.
Specifically, this is useful for small target corpora as traditional models are not typically effective in such cases. 

Considering the above motivations, we thus formalize a new problem, \textbf{Coordinated Topic Modeling} (CTM). It takes a target corpus $\mathcal{D}$, 
a reference representation $\beta^r$ for $k$ well-defined topics with their surface names $\mathcal{C}$.
As output, it aims to learn a target representation $\beta$ to best define $\mathcal{D}$.
In many cases, we can obtain a set of well-defined topics with their representation. E.g., there may exist some large corpora in a particular domain annotated with topics, and there are effective existing methods \cite{ramage2009labeled,ramage2011partially} to obtain the reference representations. Specifically, we use labeled LDA  \cite{ramage2009labeled} to get reference representation (more details on Section \ref{sec:ref_rep})


\input{EMNLP 2022/figures/architecture}

Some previous attempts incorporate prior knowledge into topic models to impose more interpretability. One such work takes document-level supervision by providing topic labels of all or a subset of documents \cite{mcauliffe2007supervised,ramage2009labeled}. While they improve the predictive ability of unsupervised topic models, they require a massive set of annotated documents to be effective. An alternative way is called topic-level supervision by providing seed words for each topic \cite{eshima2020keyword,harandizadeh2021keyword}. 
Although they make the topics converge toward the user's interest, the seed words should be only from the target corpus vocabulary, which may be impractical in many cases. Finally, the category-guided topic mining \cite{meng2020discriminative} considers the topic names as the only supervision for mining user-desired discriminative topics. However, it also assumes that the name of each topic appears in the corpus. Moreover, none of the above works does impose requirements of making topics comparable over multiple corpora (see Appendix \ref{sec:related_work} for a detailed discussion on related work).

From the above discussion, we identify the following two \textbf{requirements} that the solution of CTM needs to meet. \textbf{(1)}
The target representation should be learned based on reference representation by capturing the target corpus-specific aspects; \textbf{(2)} It also needs to relate the documents in the corpus with the global semantics of each topic represented by its surface name so that it maintains general interpretability and comparability. Although previous work considers mining topics from user guidance, none of those fulfill these two requirements.

There are several challenges to fulfilling the above requirements.
The \textbf{first challenge} is \textit{handling the vocabulary mismatch} problem. As the vocabulary of $\beta^{r}$ and $\mathcal{D}$ can be the different, $\beta^{r}$ cannot be directly calibrated into the target $\beta$. 
To solve this problem, we propose a method called \textit{reference projection}. Here, the main idea is first to generate a proxy representation $\tilde{\beta}^{r}$ of the target $\beta$ having the same vocabulary with reference $\beta^{r}$, making it comparable for supervision. It then enforces $\tilde{\beta}^{r}$ and $\beta^{r}$ to be as close as possible, which indirectly allows the target $\beta$ to be guided by given $\beta^{r}$ even if they have different vocabularies. Moreover, another \textbf{benefit} of $\tilde{\beta}^{r}$ is that it enables directly comparing two corpora given the same topics set.

The \textbf{second challenge} is \textit{providing surface names guidance}. We only know the surface name of each given topic without further knowledge of how each document corresponds to them. Therefore, we generate document-level supervision from the surface names using the \textit{textual entailment} approach \cite{yin2019benchmarking}. It imitates how humans determine the topic(s) of a document by filling in a template (e.g., ``this document is about <topic name>''). In this paper, we utilize a pre-trained textual entailment model \cite{liu2019roberta} with given surface names to generate document-level distribution matrix $\theta^{t}$. This $\theta^{t}$ later guides CTM by relating the global semantics of surface names with given documents.

The \textbf{final challenge} is \textit{combining two supervisions}. Now, we have topic-level supervision from prior projection and document-level supervision from the textual entailment model. To combine these two supervisions, we \textbf{propose} a method called \textit{embedding based coordinated topic model}, \textbf{ECTM}. It exploits the architecture of an embedded topic model (ETM) \cite{dieng2020topic}. The main idea is to regularize the objective of ETM using our two proposed supervisions. To further generalize ECTM, we employ a mechanism similar to \textit{self-training} \cite{meng2020text} where we iteratively use the model's current output to update $\theta^{t}$.
    
Our \textbf{contributions} can be summarized as follows. \textbf{First}, we propose \textit{coordinated topic modeling}, a new problem for modeling a text corpus using well-defined topics as reference. \textbf{Second}, we develop an embedding-based framework for solving the problem. It uses given reference representation to effectively model a corpus while maintaining the semantics of given surface names. \textbf{Third}, We propose methods for generating topic-level and document-level supervisions using an introduced projection mechanism and the textual entailment approach, respectively. We then combine these two supervisions into a unified model to solve the CTM problem. \textbf{Finally}, we conduct a comprehensive set of experiments on multiple domains for different tasks, demonstrating our framework's superiority against strongly designed baselines.






%% file: EMNLP 2022/figures/example.tex
\begin{figure}
\centering
\includegraphics[width=0.8\linewidth]{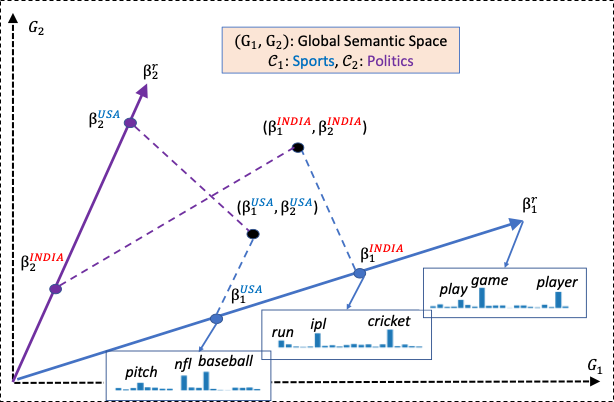}
\caption{Coordinated interpretation of Topic model}
\label{fig:example}
\vspace{-5.0mm}
\end{figure}

%% file: EMNLP 2022/figures/architecture.tex
\begin{figure*}
\centering
\includegraphics[width=1.0\linewidth]{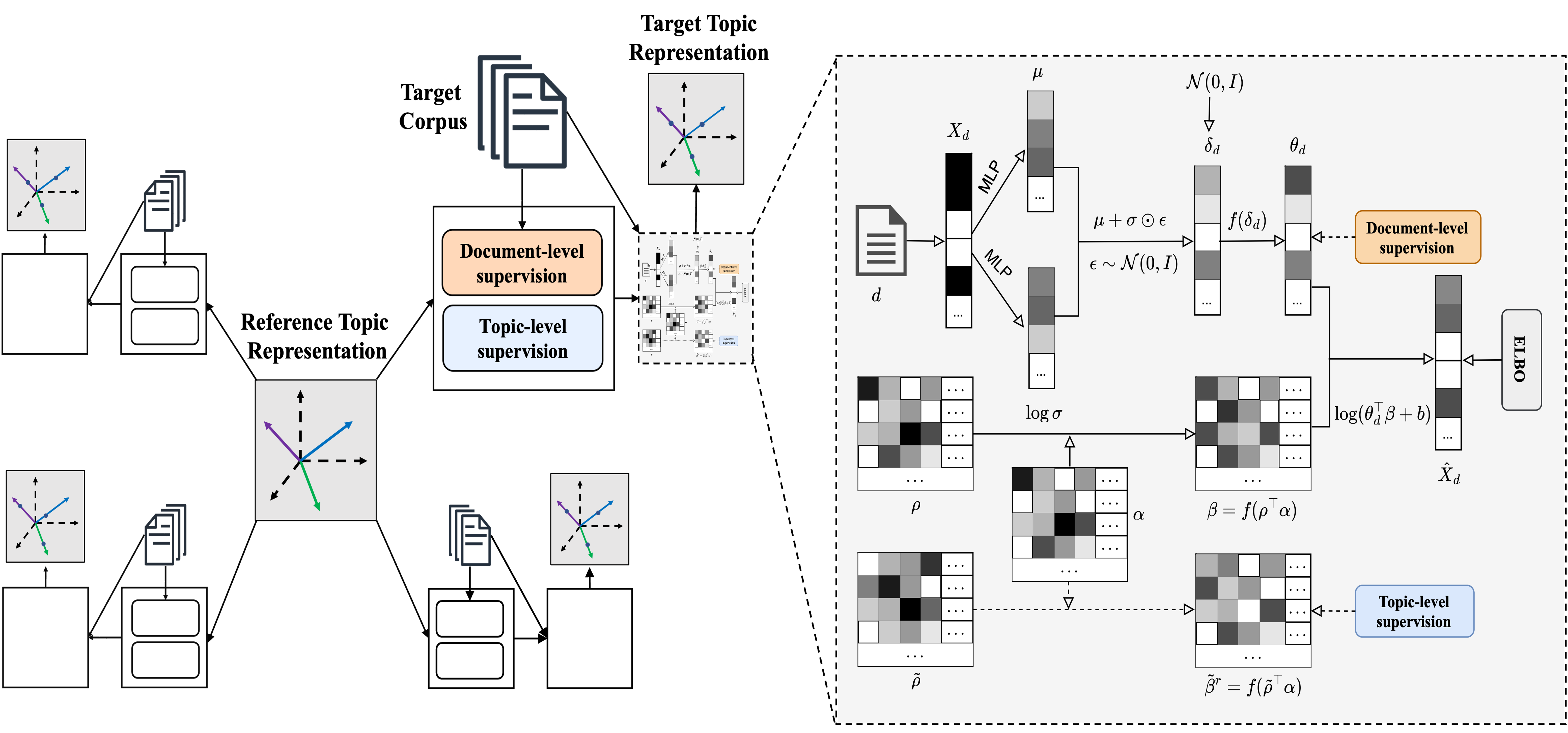}
\caption{Proposed Architecture}
\label{fig:arch}
\vspace{-5.0mm}
\end{figure*}

%% file: EMNLP 2022/sections/methodology.tex
\section{Proposed Methodology}
\label{sec:method}
{\flushleft \textbf{Overview.}} 
To solve the CTM problem, we design a method ECTM (Sec. \ref{sec:ECTM}) based on an embedded topic model (ETM) \cite{dieng2020topic} by imposing our requirements. First, we incorporate a topic-level supervision (Sec. \ref{sec:topic-level-sup}) from given reference representation $\beta^{r}$ by introducing a mechanism called \textit{reference projection}. It generates a proxy representation of intended $\beta$ with the same vocabulary dimension with $\beta^{r}$, thus enabling supervision. Second, we include a document-level supervision (Sec. \ref{sec:doc-level-sup}) from the given surface name of each topic by employing the \textit{textual entailment} approach \cite{yin2019benchmarking}. 
Finally, we combine two supervisions (Sec. \ref{sec:comb-sup}) by regularizing the ETM's objective,
To further generalize the model, we employ a self-training \cite{meng2020text} by iteratively using the model's current output to update the supervision. The rest of the section first reviews ETM and then presents our ECTM.

\subsection{Embedded Topic Model}

ETM is a neural topic model that uses vector representation of both words and topics to improve topic quality and predictive accuracy of LDA \cite{blei2003latent}. 
Let $\rho \in \mathbb{R}^{L\times V}$ and $\alpha \in \mathbb{R}^{L\times K}$ be $L$-dimensional embeddings of $V$ vocabulary words and $K$ latent topics respectively, ETM defines $k^{th}$ topic $\beta_k = f(\rho^\top \alpha_{k})$ where $f(\cdot)$ is the softmax function. It uses $\alpha$ in its generative process of $d^{th}$ document of corpus $\mathcal{D}$ as follows:
\begin{enumerate}[nolistsep,leftmargin=*]
    \item Draw topic proportion $\theta_d \sim \mathcal{LM}(0,I)$ where $\mathcal{LM}(\cdot)$ is a logistic normal distribution \cite{atchison1980logistic} that transforms a standard Gaussian random variable to the simplex. In other word, $\theta_d = f(\delta_d)$ where $\delta_d = \mathcal{N}(0,I)$.
    \item For each word $n$ in the document:
    \begin{enumerate}
        \item Draw topic assignment $z_{dn} \sim Cat(\theta_d)$.
        \item Draw the word $w_{dn}\sim$ $f(\rho^\top$
        $ \alpha_{z_{dn}})$.
    \end{enumerate}
\end{enumerate}

Here, $Cat(\cdot)$ denotes the categorical distribution. ETM employs variation inference \cite{jordan1999introduction,blei2017variational} that uses the evidence lower bound (ELBO) as a function of the model parameters ($\alpha,\rho$) and the variational parameters $v$:

\resizebox{0.9\linewidth}{!}{
\begin{minipage}{\linewidth}
\vspace{-4mm}
\begin{align}
    \mathcal{L}(\Theta) = \sum_{d\in \mathcal{D}} \sum_{n=1}^{N_d} \mathbb{E}_q[\log p(w_{dn} \mid \delta_d, \rho, \alpha)] - \nonumber \\
    \sum_{d\in \mathcal{D}} KL (q(\delta_d;w_d, v) \mid \mid p(\delta_d)),
    \label{eq:elbo}
\end{align}
  \end{minipage}
}
where $\Theta$ represents the model and variational parameters. $q(\cdot)$ is a Gaussian whose mean, and variance are estimated from a neural network.

As a function of model parameters, ELBO in Eq. \eqref{eq:elbo} is equivalent to the expected complete log-likelihood maximization. On the other hand, the first term in Eq. \eqref{eq:elbo} encourages variational parameters to place mass on topic proportions $\delta_d$ that explains the observed words, and the second term forces them to be as close as possible to $p(\delta_d)$.

\subsection{ECTM}
\label{sec:ECTM}
The proposed ECTM uses ETM as the base model to solve the CTM problem. There are several reasons for this. 
Firstly, ETM combines the strength of the neural topic model and word embedding for modeling corpus more effectively. 
Secondly, it lets us use pre-trained word embedding to map words in a common vector space even if the words do not appear in the target corpus vocabulary. Thirdly, we can regularize the objectives of ETM by imposing our problem-specific requirements. 

Figure \ref{fig:arch} shows the base model, excluding the parts involving dashed lines. Similar to ETM, we use a two-layer perceptron to encode the bag of words (BOW) representation $X_d$ of a document $d$ into document-topic distribution $\theta_d$. At the same time, vector representation of words $\rho$ and topics $\alpha$  generate topic-word distribution $\beta$. Finally, from $\theta_d$ and $\beta$, $X_d$ is reconstructed as $\hat{X}_d = \log (\theta_d^\top \beta + b)$.
where $b$ is the \textit{background bias} representing the relative frequency of the vocabulary words in $\mathcal{D}$. Unlike ETM, we include $b$ to account for words with approximately the same frequency across documents. It helps produce coherent topics by weighing down common words. Based upon this base model, ECTM consists of the following three components for incorporating the CTM's requirements.




\subsubsection{Topic-level Supervision}
\label{sec:topic-level-sup}
Our first requirement stems from the fact that, in many cases, describing a corpus with some arbitrary topics incurs a burden for users to understand them first. Instead, it is more convenient to summarize the corpus over some well-defined topics. Thus, in CTM, we consider a set of well-defined topics with reference representation $\beta^{r}$ (i.e., word distribution) like the axes in semantic space by employing supervision to generate a target representation $\beta$ that best describes the given corpus $\mathcal{D}$. 
Here, $\beta^{r}$ may come from existing sources. E.g., a large corpus in a similar domain annotated with topics can be used by supervised topic models as they are effective for finding high-quality topics from a large annotated corpus. 

Now, the question is, how can we use $\beta^{r}$ as reference axes in the semantic space to guide generating $\beta$? One possible solution can be constraining the ETM so that the generated $\beta$ comes close to $\beta^{r}$ while also maximizing ELBO. However, the problem is that we cannot assume that $\beta^{r}$ and $\beta$ share the same vocabulary. Hence, $\beta^{r}$ cannot be used directly as guidance. To solve this problem, we take an indirect way of providing supervision which we call \textit{reference projection}.

{\flushleft \textbf{Reference Projection.}} 
As shown in Figure \ref{fig:arch}, alongside $\beta$, we also generate a representation $\tilde{\beta}^{r} = f(\tilde{\rho}^\top \alpha)$. Here, $\tilde{\rho}$ is the embedding matrix of the vocabulary that the given reference $\beta^{r}$ is based on. Now, we aim to enforce $\tilde{\beta}^{r}$ and $\beta^{r}$ to be as close as possible by minimizing the following:
\vspace{-4mm}
\begin{align}
    R_{\beta} = \frac{1}{k}\sum_{j = 1}^{k} KL(\beta^{r}_j, \tilde{\beta}^{r}_j).
\end{align}
where $k$ is the number of topics.
Here, minimizing $R_\beta$ makes the model update the topic embedding matrix $\alpha$, which involves generating $\beta$. Thus, it indirectly guides $\beta$ even if there is a vocabulary mismatch between $\beta$ and $\beta^{r}$. In other words, $\beta$ is a kind of projection of $\beta^{r}$ on the target corpus vocabulary dimension.

\subsubsection{Document-level Supervision}
\label{sec:doc-level-sup}

\input{EMNLP 2022/figures/doc_supervision}

According to the second requirement, we want to maintain a global semantic of the given topic represented by its surface name. The reason is that if the target representation of each topic deviates much from its well-defined meaning, it may compromise the interpretability. 

For this, we first try to understand the semantics of a document by relating it with given surface names. Mainly, we use a textual entailment approach \cite{yin2019benchmarking} to calculate the probability of a document belonging to a topic (i.e., surface name). As humans, based on a document's content, we can determine the topic(s) by asking to what extent that document belongs to a topic out of some options. Similarly, using the textual entailment approach, we imitate humans to determine the probability that a document is about a topic by creating a hypothesis by filling in the surface name in a template (e.g., ``this document is about <surface name>''), given the context document.

Given the surface names $\mathcal{C}$, we generate $\theta^{t}$ for document-level supervision by using a pre-trained textual entailment model \cite{liu2019roberta} fine-tuned on Yahoo Answers topic classification\footnote{\url{https://huggingface.co/joeddav/bart-large-mnli-yahoo-answers}}. It considers an input document $d$ as the ``premise'', creates a ``hypothesis'' made of a template filled with a $c_k \in \mathcal{C}$, and generates a probability $p_{dk}$ denoting to what extent the premise entails the hypothesis. We have two major choices of $\theta^{t}$ from these generated probabilities. Firstly, we can use hard  labeling \cite{lee2013pseudo} which converts high-probability over a threshold $\tau$ to one-hot labels, i.e., $\theta_{dk}^{t} = 1(p_{dk} > \tau)$ where $1(\cdot)$ is the indicator function. Secondly, we can use the generated $p_{dk}$ as it is for $\theta_{dk}^{t}$ as a soft label. In soft labeling, we can further use an approach \cite{bhatia2016automatic} which elevates the high-probability label while demoting the low-probability ones. More specifically, it squares and normalizes the $p_{dk}$ as follows: 
\begin{align}
    \theta^{t}_{dk} = \frac{p_{dk}^2 / f_k}{\sum_{k^\prime} p_{d{k^\prime}}^2 / f_{k^\prime}}, \quad f_k = \sum_{d \in \mathcal{D}} p_{dk}.
\end{align}

We observe that the soft labeling strategy consistently performs better and provides more stable results than the hard labeling. The likely reason is that as hard labeling considers high-probability topics as direct labels, it is more susceptible to error propagation. Moreover, another drawback of hard labeling is that we need to set a threshold explicitly, while soft labeling does not require that. Figure \ref{fig:doc_sup} shows the overview of generating $\theta^{t}$. 

The $\theta^{t}$ is then used to provide document-level supervision by minimizing the following :
\begin{align}
    R_{\theta} = \frac{1}{\mid \mathcal{D} \mid}\sum_{d\in \mathcal{D}} KL(\theta^{t}_d, \theta_d).
\end{align}

As we use an existing knowledge source pre-trained on a massive volume of documents over various domains, this supervision helps maintain the global semantics of each topic in its target representation. Moreover, it also improves the model's predictive power to identify its topic. 

\subsubsection{Unification and Self-training}
\label{sec:comb-sup}
Now, we have topic-level and document-level supervision $R_{\beta}$ and $R_{\theta}$ respectively by imposing CTM's requirements. We unify them into one model by constraining the objective of our base model as follows:
\begin{align}
    \mathcal{L}(\Theta) = ELBO - \lambda_{\beta} R_{\beta} - \lambda_{\theta} R_{\theta},
    \label{eq:objective}
\end{align}
where $\lambda_{\beta}$ and $\lambda_{\theta}$ are the regularization weights for $R_{\beta}$ and $R_{\theta}$ respectively. Maximizing Eq. \eqref{eq:objective} jointly ensures the following objectives: (1) The ELBO part enforces the model to explain $\mathcal{D}$ by reducing the reconstruction error; (2) $R_{\beta}$ encourages the model to converge $\beta$ in the direction of $\beta^{r}$; and (3) $R_{\theta}$ encourages the model to maintain the global semantics of given coordinates in $\beta$ by enforcing $\theta$ and $\theta^{t}$ as close as possible.

Finally, to further generalize the model's predictive strength in $\theta$, we use a \textit{self-training} \cite{meng2020text} like mechanism. Here, the main idea is to iteratively use the model's current $\theta$ to update $\theta^{t}$ for supervision. More specifically, we update $\theta^{t}$ after every $50$ iteration as follows:
\begin{align}
    \theta^{t} = 0.5*\theta^{t} + 0.5*\theta.
    \label{eq:prior_update}
\end{align}
It also reduces the chance of error propagation from the textual entailment model.
We summarize our ECTM framework in Algorithm \ref{alg:algorithm}.

\input{EMNLP 2022/sections/algorithm}

%% file: EMNLP 2022/figures/doc_supervision.tex
\begin{figure}
\centering
\includegraphics[width=0.9\linewidth]{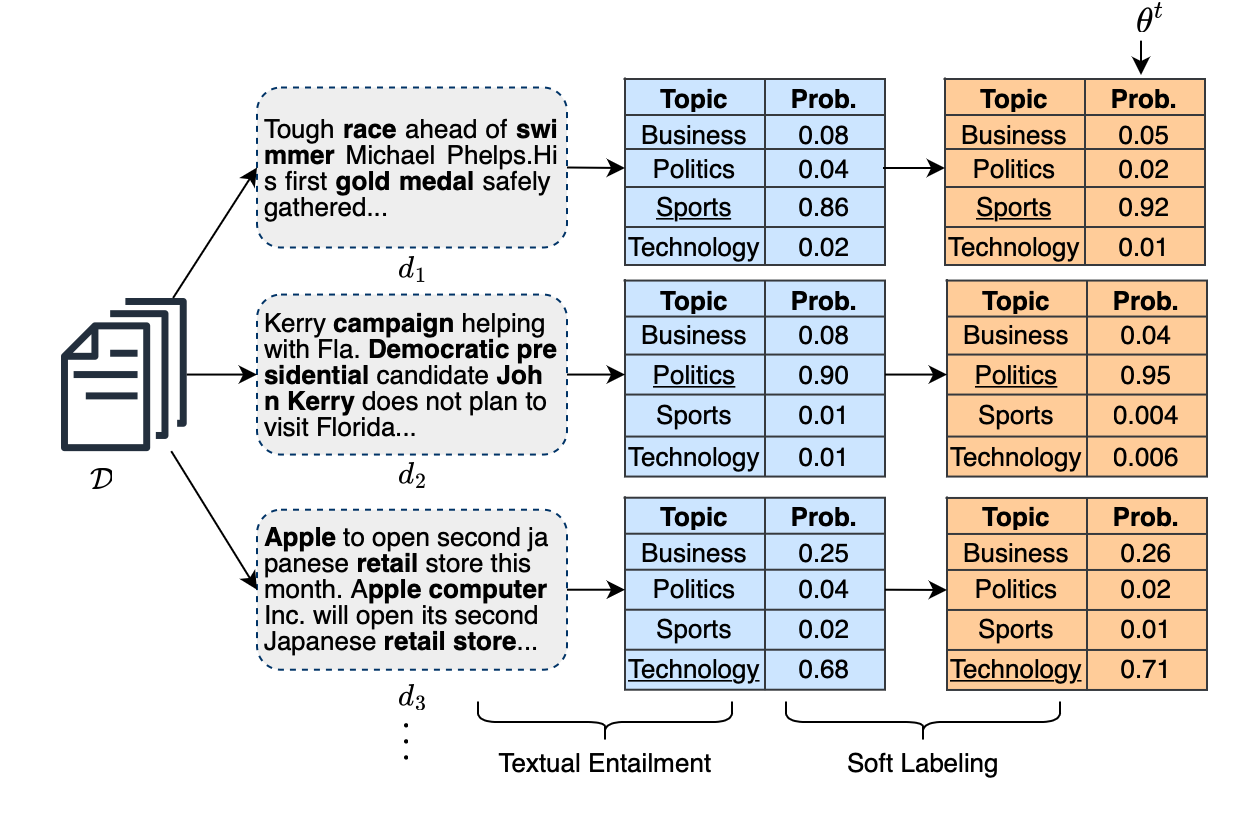}
\vspace{-3.0mm}
\caption{Document-level Supervision Generation}
\label{fig:doc_sup}
\vspace{-5.0mm}
\end{figure}

%% file: EMNLP 2022/sections/algorithm.tex
\begin{algorithm}[tb]
\caption{ECTM}
\label{alg:algorithm}
\small
\textbf{Input}: A target corpus $\mathcal{D}$; a set of topics with surface names $\mathcal{C}$ and reference representation $\beta^{r}$; a pre-trained word embedding $\mathcal{W}$; an entailment model $\mathcal{E}$. \\
\textbf{Output}: Trained Model $\mathcal{M}$ with $\beta$.\\
\begin{algorithmic}[1]
    \STATE Use $\mathcal{E}$ to initialize $\theta^{t}$
    $\gets$ Section \ref{sec:doc-level-sup};
    \STATE From $\mathcal{W}$, obtain $\rho$ and $\tilde{\rho}$;
    \STATE $B$ $\gets$ Total number of batches;
    \FOR{$i \gets$ 0 to $B-1$} 
    \STATE Train model $\mathcal{M}$ on $\mathcal{D}$ with Eq. \eqref{eq:objective};
    \IF{$i$ mod $50 = 0$}
        \STATE Update $\theta^{t}$ with Eq. \eqref{eq:prior_update};
    \ENDIF
    \ENDFOR
    \STATE return $\mathcal{M}$;
\end{algorithmic}
\end{algorithm}

%% file: EMNLP 2022/sections/experiments.tex
\section{Experiments}
\label{sec:experiments}
In this section, we employ empirical evaluations, which are designed mainly to answer the following research questions (RQs):
\textbf{RQ1.} How effective is the ECTM quantitatively in terms of the quality of topics for the CTM problem and text classification performance?
\textbf{RQ2.} How does the ECTM qualitatively perform in terms of interpretability and distinctiveness of generated topic words?
\textbf{RQ3.} How does ECTM benefit the task of comparing multiple corpora?
\textbf{RQ4.} How does each part of ECTM contribute to its performance?
 
 
 


\subsection{Experiment Setup}

{\flushleft \textbf{Datasets.}} We use datasets from three different domains \textit{news articles}, \textit{review sentiment} and \textit{academic articles} domains. For news articles, we obtain three datasets: (1) 20 Newsgroup corpus\footnote{\url{http://qwone.com/~jason/20Newsgroups/}} (\textbf{20Newsg}); (2) New York Times annotated corpus (\textbf{NYT}) \cite{sandhaus2008new}; (3) AG’s News dataset (\textbf{AGNews}) from \cite{yang2016hierarchical}. For review sentiment domain, we use: (1) Yelp restaurant review dataset used in \cite{meng2020discriminative} (\textbf{Yelp-Sent}); (2) IMDB Movie Review dataset used in \cite{hoang2019towards} (\textbf{IMDB-Sent}). Finally for academic articles: (1) Arxiv Artificial Intelligence (AI) article abstracts spanning 2020-2022 \footnote{\href{https://www.kaggle.com/Cornell-University/arxiv}{https://www.kaggle.com/Cornell-University/arxiv}} (\textbf{Arxiv-AI}); (2) Microsoft Academic Graph AI article abstracts \cite{sinha2015overview} (\textbf{MAG-AI}). 

{\flushleft \textbf{Baselines.}}
We compare our model with the following baselines.
\begin{itemize}[nolistsep,leftmargin=*]
\item GLDA: Guided LDA \cite{jagarlamudi2012incorporating} biases LDA's generative process using topic-level priors over vocabulary from given seed words.
\item Sup+LLDA: Supervised Labeled LDA is based on Labeled-LDA \cite{ramage2009labeled} where a label for each document is predicted from a supervised BERT  classifier \footnote{\url{https://huggingface.co/docs/transformers/tasks/sequence_classification}} learned on annotated reference corpus.

\item ZS+LLDA: Zero-Shot Labeled LDA is also based on Labeled-LDA \cite{ramage2009labeled} where a label for each document is inferred from given surface names using a Zero-Shot classification \cite{yin2019benchmarking}.


\item ACorEx: Unlike LDA, Anchored CorEx \cite{gallagher2017anchored} is not based on generative assumptions and uses topic correlation to learn maximally informative topics. Moreover, it uses user-provided seed words as anchors to bias compression of the original corpus.

\item AVIAD: AVIAD \cite{hoang2019towards} extends the Autoencoding Variational Inference for Topic Models (AVITM) \cite{srivastava2017autoencoding} approach to incorporate prior knowledge from seed words by modifying the loss function to infer desired topics. It uses a variational autoencoder like ETM but does not employ word embedding in the modeling.

\item KeyETM: Keyword Assisted ETM \cite{harandizadeh2021keyword} modifies ETM's objective to include prior knowledge from given seed words.
\end{itemize}

{\flushleft \textbf{Reference Topic Representation.}}
\label{sec:ref_rep}
As an input, CTM requires prior knowledge $\beta^{r}$. 
For each domain, we use a large annotated corpus as a reference to get $\beta^{r}$ using labeled LDA (LLDA) \cite{ramage2009labeled}. Because LLDA is effective in producing high-quality topics when there is a large annotated corpus. E.g., in news domain, we use a large annotated corpus \textbf{AGNews} as reference with topic names \textit{business}, \textit{politics}, \textit{sports} and \textit{technology}. In the sentiment domain, the reference corpus we used is \textbf{IMDB} with positive (\textit{good}) and negative (\textit{bad}) sentiments. Finally, we use the \textbf{MAG-AI} as reference corpus for the AI domain to generate prior, annotated with four topics: \textit{computer vision} (CV), \textit{information retrieval} (IR), \textit{machine learning} (ML), and \textit{natural language processing} (NLP). 

{\flushleft \textbf{Seed words.}}
For the baselines that require seed words, we collect them from given surface names $\mathcal{C}$ and top topic words based on given $\beta^{r}$. If a word does not appear in the corresponding target corpus, we replace that with the most similar word using cosine similarity. As suggested in the baselines, we provide 10 seed words for each topic.

The implementation details of our model and baselines are specified in Appendix \ref{sec:implement}.

\input{EMNLP 2022/tables/topic_quality}
\input{EMNLP 2022/tables/top_words}

\subsection{Topic Quality}

{\flushleft \textbf{Evaluation Metrics.}}
 We use the following three well-defined quantitative measurements to evaluate the inferred topics' quality:

\begin{itemize}[nolistsep,leftmargin=*]
 \item Topic coherence (TC) is a standard measure of interpretability based on the average point-wise mutual information between randomly drawn two words from the same document \cite{lau2014machine}.
 \item Topic diversity (TD) measures the percentage of unique words in the top 25 words of all topics \cite{dieng2020topic}. It captures the semantical diverseness of the inferred topics.
 \item Topic quality (TQ) \cite{dieng2020topic} is the overall metric for measuring the quality of topics as the product of topic coherence and diversity.
\end{itemize}

{\flushleft \textbf{Results and Discussions.}}
We first show the quantitative results of topic quality in Table \ref{tab:topic_quality}. 
The results suggest that, in general, ECTM generates more coherent and interpretable topics than other baselines. In some cases, other methods produce more diverse topics than ours. E.g., ECTM's generated topics' diversity scores are slightly lower than the best one in \textbf{NYT} and \textbf{Arxiv-AI} datasets. 
However, our method significantly outperforms others in quality scores in those cases by producing more coherent topics. Thus, ECTM produces more interpretable topics while also maintaining diversity.

In Table \ref{tab:qualitative}, we show randomly selected two topics from each dataset and show the top-5 words under each topic. We also show the top-5 words from the given $\beta^{r}$. Words that authors determined to be irrelevant to the corresponding topic are marked with ($\times$). Overall, our method-generated topic words are relevant and easily interpretable in nearly all cases compared to baselines. We also observe that the topics produced by AcorEx are reasonably good in terms of interpretability. 
However, AcorEx's produced topics strictly converge toward the prior representation rather than adapting according to the target corpus. E.g., in \textit{sports} topic, 3 out of 5 topic words overlap with priors topic words. 
In contrast, our method tends to capture the target corpus-specific aspects of the given topics.


On the other hand, AVIAD suffers from the opposite issue. It adapts too much that the topics become very difficult to understand. The KeyETM with a similar base model (ETM) to ours performs better when the target corpus is balanced. E.g., as \textbf{20Newsg} is a comparatively balanced dataset, the keyETM performs well.
In contrast, our proposed ECTM consistently performs better as it enjoys the benefit of both topic-level supervision from $\beta^{r}$ and document-level supervision from existing knowledge sources to make the topics adjusted to the target corpus as well as not going away from well-known semantics of given topics' names.

\input{EMNLP 2022/tables/classification}
\vspace{-1mm}
\subsection{Text Classification}
\vspace{-1mm}
Although the primary purpose of our model is not document categorization, we can use learned $\theta$ to analyze the document representation. More specifically, we can consider the topic with maximum probability $\theta_d$ as the document's label of $d$. Therefore, we can evaluate how learned topics are distinctive and informative enough to represent a document to categorize it correctly.  
Alongside the topic models, to compare text classification performance,
we also include several supervised and semi-supervised baselines: SupBERTCLs, Zero-ShotCLs \cite{yin2019benchmarking}, WeSTCLs \cite{meng2018weakly}, and LOTCls \cite{meng2020text}, with details described in Appendix \ref{sec:semi-baseline}. 

{\flushleft \textbf{Evaluation Metrics.}} 
We use accuracy (Acc.), macro-F1 (F1-Mac) and micro-F1 (F1-Mic) scores that are commonly used in classification evaluations \cite{meng2018weakly,meng2020text} as the metrics.
{\flushleft \textbf{Results and Discussions.}}
The text classification results on \textbf{20Newsg}, \textbf{NYT}, \textbf{Yelp-Senti} and \textbf{Arxiv-AI} datasets are shown in Table \ref{tab:classification}. 
We can see that ECTM mostly performs better by all three evaluation metrics than other models with large margins. 
One reason for this better result is the document-level supervision we employ from the textual entailment model. It enforces the model to generate $\theta$ with better predictive power. It can also be explained by the comparable performance with the baselines ZS+LLDA and Zero-ShotCLs models, as they also use the textual entailment approach. 
Here, we can see Sup+LLDA and SupBERTCLs perform slightly better than ours in the Yelp-Sent dataset because it is rather an easy task to learn a classifier of binary sentiment classes from supervision. However, it performs very poorly for other more hard classification datasets.

Moreover, our model performs better even when the textual entailment model fails to perform well. E.g., in \textbf{Arxiv-AI} dataset, while the ZS+LLDA and Zero-ShotCLs have less than 0.40 for all three measures, ECTM performs significantly better by attaining close or more than 0.80 scores in the evaluation measures. This is because our model does not rely solely on the textual entailment model. It self-trains itself by updating supervision iteratively by reducing the chance of error propagation from the entailment model.

\input{EMNLP 2022/figures/corpus_comparison}
\input{EMNLP 2022/tables/context_def_words}
\vspace{-1mm}
\subsection{Corpus Comparison}
This section explores how ECTM facilitates the quantitative comparison of any two corpora given the same topics. In ECTM, as described before, while incorporating topic-level supervision, we generate a proxy topic-word distribution $\tilde{\beta}^{r}$ of the target $\beta$ to make it comparable with $\beta^{r}$. As both $\beta^{r}$ and $\tilde{\beta}^{r}$ share the same vocabulary, any two corpora with the same sets of topics can be thus quantitatively compared using their generated $\tilde{\beta}^{r}$s. To investigate the comparison, we apply our ECTM to multiple corpora over different contexts using the same topics. More specifically, we use two contexts: time and location. For time context, we use NYT articles from December 2000 and September 2001 as two target corpora. For location context, we use NYT articles on USA and Britain as two target corpora. In both cases, we use the same set of topics, and the entire NYT corpus is used to obtain the reference.

\input{EMNLP 2022/tables/ablation}

Let $\tilde{\beta}^{r(1)}$ and $\tilde{\beta}^{r(2)}$ are generated by two corpora $\mathcal{D}_1$ and $\mathcal{D}_2$ respectively, we compare $\mathcal{D}_1$ and $\mathcal{D}_2$ by calculating $ KL (\tilde{\beta}^{r(1)}_j || \tilde{\beta}^{r(2)}_j)$ for each topic $j\in \{1,\cdots, k\}$. Figure \ref{fig:corpus_comp} illustrates such results for our two contexts. Here, we see in both contexts, the topic \textit{politics} has made the highest difference between the two corresponding corpora. 
We also show some context defining words out of top 20-words obtained from ECTM generated $\beta$ for politics in Table \ref{tab:context_words} (full list is in Appendix \ref{sec:additional-corpus_comp}). 
As December 2000 was the month when  the \textit{Bush vs. Gore 2000 presidential race} was settled, the words like \textit{election}, \textit{vote} or \textit{judge} make sense. And words like \textit{attack}, \textit{terrorism} or \textit{war} are dominated in \textit{politics} topic on September 2001 as there was a terror attack that month. Similarly, the words that mainly make the difference in politics for USA and Britain are self-explanatory.

\input{EMNLP 2022/figures/param_effect}

\subsection{Ablation Study}
\label{sec:ablation}

In this section, we investigate the role of different parts of our proposed model in \textbf{20Newsg} dataset (shown in Table \ref{tab:ablation}). First, when we exclude the background bias $b$ from the model, we see the common words (i.e., \textit{time}, \textit{year}) are being dominated in the top-5 topic words, thus making it less interpretable and distinctive. It explains the effectiveness of background bias in the model. Second, when only the document-level supervision (ECTM -$R_{\beta}$) is employed, the model's predictive power gets improved. However, the topic quality scores are significantly downgraded, which is also reflected in irrelevant words (i.e., general words like \textit{great, la}) in the top-5 topic words. It justifies the intuition of our topic-level supervision to make topics interpretable by calibrating them from well-defined topics. On the other hand, only using topic-level supervision degrades the model's predictive power. Finally, the complete model balances them by generating interpretable and discriminatory topics, which is reflected in the quantitative measures.

Moreover, in Figure \ref{fig:param_effect}, we show the effect of two supervisions on the topic's adaptability in the target corpus. The vertical axis shows the divergence of learned topics with the given reference topics. If the divergence score increases, it means topics are getting more adapted to the corpus than being similar to the reference one. Figure \ref{fig:param_effect} (a) shows that if we increase the regularization weight parameter $\lambda_{\theta}$ for document-level supervision by fixing $\lambda_{\beta}$, the topics get deviated from reference topic focusing more on the target corpus content. On the other hand, while increasing topic-level regularization weight $\lambda_{\beta}$ by fixing $\lambda_{\theta}$, the model tends to converge towards the reference topics. It is intuitive and reasonable according to our discussion of model requirements.

%% file: EMNLP 2022/tables/topic_quality.tex
\begin{table}[!tb]
\centering
\resizebox{\linewidth}{!}{%
\setlength\tabcolsep{2pt}
\begin{tabular}{@{}c|ccc|ccc|ccc|ccc@{}}
\toprule
\multirow{2}{*}{Methods} & \multicolumn{3}{c|}{\textbf{20Newsg}}     & \multicolumn{3}{c|}{\textbf{NYT}}         & \multicolumn{3}{c|}{\textbf{Yelp-Senti}}  & \multicolumn{3}{c}{\textbf{Arxiv-AI}}    \\  
                         & TC & TD & TQ & TC & TD & TQ & TC & TD & TQ & TC & TD & TQ \\ \midrule
                         
GLDA                     & 0.25      & 0.87      & 0.22    & 0.26      & 0.85      & 0.22    & 0.08      & 0.80      & 0.06    & 0.09      & 0.93      & 0.09    \\
Sup+LLDA                    & 0.23      & 0.79      & 0.18    & 0.20      & 0.63      & 0.12    & 0.06      & 0.70      & 0.04    & 0.04      & 0.46      & 0.02    \\

ZS+LLDA                    & 0.23      & 0.80      & 0.18    & 0.17      & 0.65      & 0.11    & 0.06      & 0.76      & 0.05    & 0.14      & 0.80      & 0.11    \\

ACorEX                   & 0.25      & \textbf{1.00}       & 0.25    & 0.27      & \textbf{1.00}       & 0.27    & 0.07      & \textbf{1.00}       & 0.07    & -0.03      & 0.96       & -0.03    \\
AVIAD                    & 0.13      & \textbf{1.00}       & 0.13    & -0.26     & \textbf{1.00}       & -0.26   & -0.01      & \textbf{1.00}       & -0.01    & -0.34     & \textbf{1.00}       & -0.34   \\
KeyETM                   & 0.26      & \textbf{1.00}       & 0.26    & 0.19      & 0.89      & 0.17    & 0.07      & 0.92      & 0.07    & 0.04      & \textbf{1.00}       & 0.04    \\ \midrule
ECTM                     & \textbf{0.30}      & \textbf{1.00}       & \textbf{0.30}    & \textbf{0.28}      & 0.97      & \textbf{0.27}    & \textbf{0.09}      & \textbf{1.00}       & \textbf{0.09}    & \textbf{0.15}      & 0.97       & \textbf{0.15}    \\ \bottomrule
\end{tabular}}
\vspace{-2.0mm}
\caption{Quality Measures of Topic}
\label{tab:topic_quality}
\vspace{-5.00mm}
\end{table}

%% file: EMNLP 2022/tables/top_words.tex

\begin{table*}[]
\centering
\resizebox{0.8\linewidth}{!}{%
\begin{tabular}{@{}c|cc|cc|cc|cc@{}}
\toprule
\multirow{2}{*}{} & \multicolumn{2}{c|}{\textbf{20Newsg}}                                                                                                                                                                    & \multicolumn{2}{c|}{\textbf{NYT}}                                                                                                                                                                                     & \multicolumn{2}{c|}{\textbf{Yelp-Senti}}                                                                                                                                                                                  & \multicolumn{2}{c}{\textbf{Arxiv-AI}}                                                                                                                                                                                            \\ 
                         & sports                                                                                         & politics                                                                                      & business                                                                                       & technology                                                                                                 & good                                                                                                    & bad                                                                                                   & ML                                                                                                                   & IR                                                                                               \\ \toprule
\begin{tabular}[c]{@{}c@{}} Reference\\Topic\\ Words\end{tabular}                    & \begin{tabular}[c]{@{}c@{}}night\\play\\sport\\player\\beat\end{tabular}                 & \begin{tabular}[c]{@{}c@{}}leader\\ election \\ attack\\ afp\\ iraqi \end{tabular}       & \begin{tabular}[c]{@{}c@{}}stock\\ sale\\ share\\ billion\\ fall\end{tabular}           & \begin{tabular}[c]{@{}c@{}}software\\ technology\\ service\\ internet\\ launch\end{tabular}   & \begin{tabular}[c]{@{}c@{}}song\\ music\\ musical\\ wonderful\\ dance\end{tabular}                 & \begin{tabular}[c]{@{}c@{}}waste\\ awful\\ terrible\\ boring\\ poor\end{tabular}     & \begin{tabular}[c]{@{}c@{}}machine\\ learning\\ algorithm\\ optimization\\ problem\end{tabular}       & \begin{tabular}[c]{@{}c@{}}retrieval\\ document\\ query\\ search\\ base\end{tabular}       \\ \toprule
GLDA                     & \begin{tabular}[c]{@{}c@{}}game\\ team\\ year $(\times)$\\ play\\ player\end{tabular}                 & \begin{tabular}[c]{@{}c@{}}people\\ time $(\times)$\\ government\\ gun\\ year $(\times)$\end{tabular}       & \begin{tabular}[c]{@{}c@{}}company\\ percent\\ year $(\times)$\\ bank\\ market\end{tabular}           & \begin{tabular}[c]{@{}c@{}}president $(\times)$\\ bush $(\times)$\\ official $(\times)$\\ united $(\times)$\\ house $(\times)$\end{tabular}   & \begin{tabular}[c]{@{}c@{}}good\\ place $(\times)$\\ food $(\times)$\\ great\\ order $(\times)$\end{tabular}                 & \begin{tabular}[c]{@{}c@{}}order $(\times)$\\ food $(\times)$\\ time $(\times)$\\ place $(\times)$\\ service $(\times)$\end{tabular}     & \begin{tabular}[c]{@{}c@{}}adversarial\\ distribution $(\times)$\\ class $(\times)$\\ function $(\times)$\\ attack $(\times)$\end{tabular}       & \begin{tabular}[c]{@{}c@{}}graph\\ search\\ user\\ class $(\times)$\\ recommendation\end{tabular}       \\ \midrule

Sup+LLDA                   & 

\begin{tabular}[c]{@{}c@{}}game\\ team\\ year $(\times)$\\ play\\ time $(\times)$\end{tabular}                 & 

\begin{tabular}[c]{@{}c@{}} people \\ government \\ kill \\ time $(\times)$ \\ year $(\times)$ \end{tabular} & 

\begin{tabular}[c]{@{}c@{}}year $(\times)$  \\ percent \\ company \\ market \\ government $(\times)$ \end{tabular}       & 

\begin{tabular}[c]{@{}c@{}}year $(\times)$ \\ time $(\times)$ \\ people $(\times)$ \\ president $(\times)$ \\ official $(\times)$\end{tabular}      & 

\begin{tabular}[c]{@{}c@{}} place $(\times)$ \\ food $(\times)$ \\ good \\ great \\ service $(\times)$ \end{tabular} &

\begin{tabular}[c]{@{}c@{}} food $(\times)$ \\ order $(\times)$ \\ place $(\times)$ \\ service $(\times)$ \\ time $(\times)$ \end{tabular}               & 

\begin{tabular}[c]{@{}c@{}}demonstrate $(\times)$ \\ problem $(\times)$ \\ feature \\ training \\ neural \end{tabular}          & 

\begin{tabular}[c]{@{}c@{}} retrieval \\ exist $(\times)$ \\ demonstrate $(\times)$ \\ representation \\ feature $(\times)$ \end{tabular}

\\ \midrule
ZS+LLDA                   & \begin{tabular}[c]{@{}c@{}}game\\ team\\ year $(\times)$\\ play\\ player\end{tabular}                 & \begin{tabular}[c]{@{}c@{}}people\\ time $(\times)$\\ government\\ year $(\times)$\\ point $(\times)$\end{tabular} & \begin{tabular}[c]{@{}c@{}}year $(\times)$\\ percent\\ company\\ market\\ lead $(\times)$\end{tabular}       & \begin{tabular}[c]{@{}c@{}}year $(\times)$\\ time $(\times)$\\ american $(\times)$\\ official $(\times)$\\ today $(\times)$\end{tabular}      & \begin{tabular}[c]{@{}c@{}}food $(\times)$\\ place $(\times)$\\ good\\ great\\ service $(\times)$\end{tabular}               & \begin{tabular}[c]{@{}c@{}}food $(\times)$\\ order $(\times)$\\ place $(\times)$\\ service $(\times)$\\ time $(\times)$\end{tabular}     & \begin{tabular}[c]{@{}c@{}}efficient $(\times)$\\ reduce $(\times)$\\ number $(\times)$\\ leverage $(\times)$\\ module $(\times)$\end{tabular}          & \begin{tabular}[c]{@{}c@{}}retrieval\\ search\\ user\\ document\\ query\end{tabular}             \\ \midrule
ACorEX                   & \begin{tabular}[c]{@{}c@{}}point\\ play\\ player\\ league\\ beat\end{tabular}                  & \begin{tabular}[c]{@{}c@{}}force\\ country\\ attack\\ military\\ political\end{tabular}       & \begin{tabular}[c]{@{}c@{}}billion\\ business\\ buy\\ stock\\ profit\end{tabular}              & \begin{tabular}[c]{@{}c@{}}release $(\times)$\\ technology\\ phone\\ time $(\times)$\\ space\end{tabular}                & \begin{tabular}[c]{@{}c@{}}good\\ hear $(\times)$\\ beautiful\\ music $(\times)$\\ sound $(\times)$\end{tabular}             & \begin{tabular}[c]{@{}c@{}}bad\\ money $(\times)$\\ terrible\\ poor\\ waste\end{tabular}                     & \begin{tabular}[c]{@{}c@{}}optimization\\ gradient\\ convergence\\ stochastic\\ print $(\times)$\end{tabular}               & \begin{tabular}[c]{@{}c@{}}search\\ document\\ query\\ retrieval\\ semantics\end{tabular}        \\ \midrule
AVIAD                    & \begin{tabular}[c]{@{}c@{}}robitaille $(\times)$\\ probert $(\times)$\\ howe $(\times)$\\ player\\ nhl\end{tabular} & \begin{tabular}[c]{@{}c@{}}tragedy $(\times)$\\ policy\\ serbian\\ freedom\\ unite $(\times)$\end{tabular}  & \begin{tabular}[c]{@{}c@{}}sanwa $(\times)$\\ zoete $(\times)$\\ earning\\ overprice\\ acquirer\end{tabular} & \begin{tabular}[c]{@{}c@{}}genscher $(\times)$\\ enlargement $(\times)$\\ abm $(\times)$\\ teng $(\times)$\\ chechnya $(\times)$\end{tabular} & \begin{tabular}[c]{@{}c@{}}traditional $(\times)$\\ snow $(\times)$\\ filling\\ bisque $(\times)$\\ seaweed $(\times)$\end{tabular} & \begin{tabular}[c]{@{}c@{}}email $(\times)$\\ upset\\ management $(\times)$\\ yell\\ acknowledge $(\times)$\end{tabular}   & \begin{tabular}[c]{@{}c@{}}bind\\ analytically $(\times)$\\ certify $(\times)$\\ arm $(\times)$\\ pruning\end{tabular}                    & \begin{tabular}[c]{@{}c@{}}ehr\\ healthy $(\times)$\\ progression $(\times)$\\ patient $(\times)$\\ ehrs\end{tabular} \\ \midrule
KeyETM                   & \begin{tabular}[c]{@{}c@{}}game\\ team\\ season\\ play\\ win\end{tabular}                      & \begin{tabular}[c]{@{}c@{}}people\\ government\\ person $(\times)$\\ armenian\\ law\end{tabular}     & \begin{tabular}[c]{@{}c@{}}year $(\times)$\\ percent\\ market\\ time $(\times)$\\ month $(\times)$\end{tabular}     & \begin{tabular}[c]{@{}c@{}}company $(\times)$\\ bank $(\times)$\\ japan $(\times)$\\ china $(\times)$\\ russia $(\times)$\end{tabular}        & \begin{tabular}[c]{@{}c@{}}good\\ place $(\times)$\\ great\\ time $(\times)$\\ love\end{tabular}                      & \begin{tabular}[c]{@{}c@{}}food $(\times)$\\ order $(\times)$\\ service $(\times)$\\ eat $(\times)$\\ restaurant $(\times)$\end{tabular} & \begin{tabular}[c]{@{}c@{}}function\\ estimation $(\times)$\\ distribution $(\times)$\\ parameter $(\times)$\\ efficient $(\times)$\end{tabular} & \begin{tabular}[c]{@{}c@{}}translation $(\times)$\\ user\\ search\\ annotation\\ point $(\times)$\end{tabular} \\ \midrule
ECTM                     & \begin{tabular}[c]{@{}c@{}}game\\ team\\ win\\ season\\ league\end{tabular}                    & \begin{tabular}[c]{@{}c@{}}government\\ war\\ military\\ armenian\\ attack\end{tabular}       & \begin{tabular}[c]{@{}c@{}}company\\ bank\\ percent\\ market\\ price\end{tabular}              & \begin{tabular}[c]{@{}c@{}}space\\ site\\ technology\\ station\\ network\end{tabular}                      & \begin{tabular}[c]{@{}c@{}}great\\ music $(\times)$\\ love\\ wonderful\\ amazing\end{tabular}                  & \begin{tabular}[c]{@{}c@{}}waste\\ awful\\ terrible\\ bad\\ horrible\end{tabular}                     & \begin{tabular}[c]{@{}c@{}}optimization\\ convergence\\ stochastic\\ gradient\\ function\end{tabular}                & \begin{tabular}[c]{@{}c@{}}retrieval\\ document\\ query\\ search\\ user\end{tabular}             \\ \bottomrule
\end{tabular}
}
\vspace{-2.0mm}
\caption{Qualitative Evaluation}
\label{tab:qualitative}
\vspace{-5.00mm}
\end{table*}

%% file: EMNLP 2022/tables/classification.tex
\begin{table}[!tb]
\centering
\resizebox{\linewidth}{!}{%
\setlength\tabcolsep{1.5pt}
\begin{tabular}{@{}c|ccc|ccc|ccc|ccc@{}}
\toprule
\multirow{2}{*}{Methods} & \multicolumn{3}{c|}{\textbf{20Newsg}}     & \multicolumn{3}{c|}{\textbf{NYT}}         & \multicolumn{3}{c|}{\textbf{Yelp-Senti}}  & \multicolumn{3}{c}{\textbf{Arxiv-AI}}           \\  
                        & Acc. & F1-Mac & F1-Mic & Acc. & F1-Mac & F1-Mic & Acc. & F1-Mac & F1-Mic & Acc. & F1-Mac & F1-Mic \\ \midrule
GLDA                    & 0.70     & 0.71       & 0.73          & 0.70     & 0.59       & 0.75          & 0.75     & 0.75       & 0.75          & 0.67     & 0.59       & 0.71          \\

Sup+LLDA                    &0.68  &0.53  &0.62              &0.77 &0.70 &0.80                      &\textbf{0.94} &\textbf{0.94 }&\textbf{0.94  }       &0.38 &0.37 &0.38                    \\

ZS+LLDA                   & 0.89     & 0.80       & 0.88          & 0.91     & 0.79       & 0.90          & 0.87     & 0.87       & 0.87          & 0.34     & 0.34       & 0.33          \\
ACorEX                  & 0.38     & 0.43       & 0.44          & 0.72     & 0.63       & 0.73          & 0.60     & 0.59       & 0.58          & 0.66     & 0.51       & 0.66          \\
AVIAD                   & 0.72     & 0.71       & 0.74          & 0.69     & 0.52       & 0.72          & 0.75     & 0.74       & 0.75          & 0.64     & 0.57       & 0.68          \\
KeyETM                  & 0.66     & 0.62       & 0.69          & 0.39     & 0.31       & 0.40          & 0.49     & 0.34       & 0.32          & 0.30     & 0.24       & 0.23          \\ \midrule

SupBERTCLs       &0.68  &0.53  &0.62              &0.77 &0.70 &0.80                      &\textbf{0.94} &\textbf{0.94 }&\textbf{0.94  }       &0.38 &0.37 &0.38              \\

Zero-ShotCLs       & 0.89          & 0.80           & 0.88              & 0.91         & 0.79           & 0.90              & 0.87         & 0.87           &  0.87             & 0.34         & 0.34           & 0.33              \\

WesSTCLs (Names)              & 0.55          & 0.47            &  0.57              & 0.72         & 0.65           & 0.76              & 0.80         & 0.79           & 0.79              & 0.44         & 0.44           & 0.48              \\
WesSTCLs (Seeds)              & 0.73          & 0.65            & 0.76              & 0.79         & 0.70           & 0.83              & 0.64          & 0.64            & 0.64              & 0.79         & 0.68            & 0.81              \\
LOTCLs                & 0.84          & 0.68            & 0.81               & 0.84         & 0.75           & 0.86              & 0.76          & 0.75           & 0.75              & 0.07         & 0.09           & 0.06              \\ \midrule
ECTM                    & \textbf{0.91}     & \textbf{0.85}       & \textbf{0.91}          & \textbf{0.91}     & \textbf{0.83}       & \textbf{0.92}          & 0.90     & 0.90       & 0.90          & \textbf{0.83}     & \textbf{0.76}       & \textbf{0.83}          \\ \bottomrule
\end{tabular}}
\vspace{-2.0mm}
\caption{Text Classification Performance}
\label{tab:classification}
\vspace{-5.00mm}
\end{table}

%% file: EMNLP 2022/figures/corpus_comparison.tex
\begin{figure}
\centering
\includegraphics[width=0.9\linewidth]{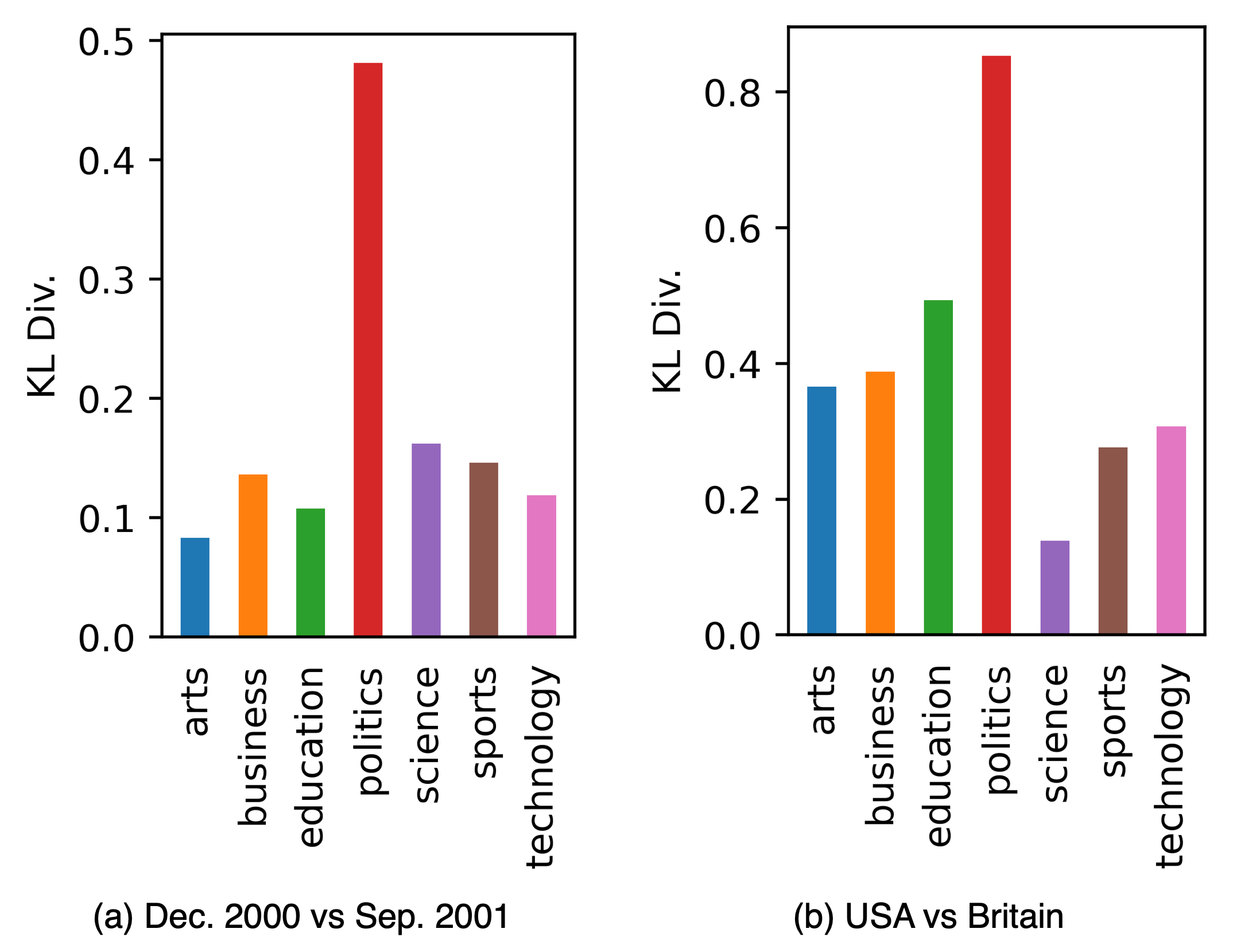}
\vspace{-1.0mm}
\caption{Corpus Comparison}
\label{fig:corpus_comp}
\end{figure}

%% file: EMNLP 2022/tables/context_def_words.tex
\begin{table}[!tb]
\resizebox{1.0\linewidth}{!}{%
\begin{tabular}{@{}l|l@{}}
\toprule
Dec. 2000 & court, election, vote, bush, president, florida, judge, supreme            \\ \midrule
Sep. 2001 & attack, war, president, military terrorism, terrorist, afghanistan, muslim \\ \midrule
USA       & republican, president, governor, senator, clinton, bush, giuliani, senate  \\ \midrule
Britain   & prime, minister, ireland, irish, northern, blair, union, thatcher          \\ \bottomrule
\end{tabular}}
\vspace{-2.0mm}
\caption{Context defining words in politics topic}
\label{tab:context_words}
\vspace{-5.0mm}
\end{table}

%% file: EMNLP 2022/tables/ablation.tex
\begin{table}[!ht]
\centering
\resizebox{1.0\linewidth}{!}{%
\begin{tabular}{@{}p{0.25\linewidth}|p{0.05\linewidth}p{0.05\linewidth}p{0.08\linewidth}|p{0.05\linewidth}p{0.05\linewidth}p{0.08\linewidth}|c@{}}

\toprule
\multirow{2}{*}{Methods}                                      & \multicolumn{3}{c|}{Quality}            & \multicolumn{3}{c|}{Classification}            & Top 5 Words of Topic                         \\
                                                              & TC     & TD    & TQ       & Acc.      & F1     & F1     & Business                            \\ \midrule
ECTM - $R_\beta$ - $R_\theta$ - $b$ & 0.24          & 0.99         & 0.24          & --            & --            & --            & time, work, year, problem, people   \\
ECTM - $b$                                                      & 0.28          & 0.98         & 0.27          & 0.90          & 0.83          & 0.90          & year, president, \textbf{price}, month, week \\
ECTM - $R_\beta$                                & 0.20          & 0.99         & 0.19          & \textbf{0.92} & \textbf{0.87} & \textbf{0.92} & fan, \textbf{sale}, la, great, \textbf{pay}           \\
ECTM - $R_\theta$                               & 0.30          & 0.99         & 0.30          & 0.83          & 0.71          & 0.82          & president, \textbf{price}, \textbf{buy}, \textbf{stock}, \textbf{oil}   \\
ECTM                                                          & \textbf{0.30} & \textbf{1.0} & \textbf{0.30} & 0.91          & 0.85          & 0.91          & \textbf{price}, \textbf{stock}, \textbf{sale}, \textbf{buy}, \textbf{oil}        \\ \bottomrule
\end{tabular}}
\caption{Ablation Study}
\label{tab:ablation}
\end{table}

%% file: EMNLP 2022/figures/param_effect.tex
\begin{figure}
\centering
\includegraphics[width=1.0\linewidth]{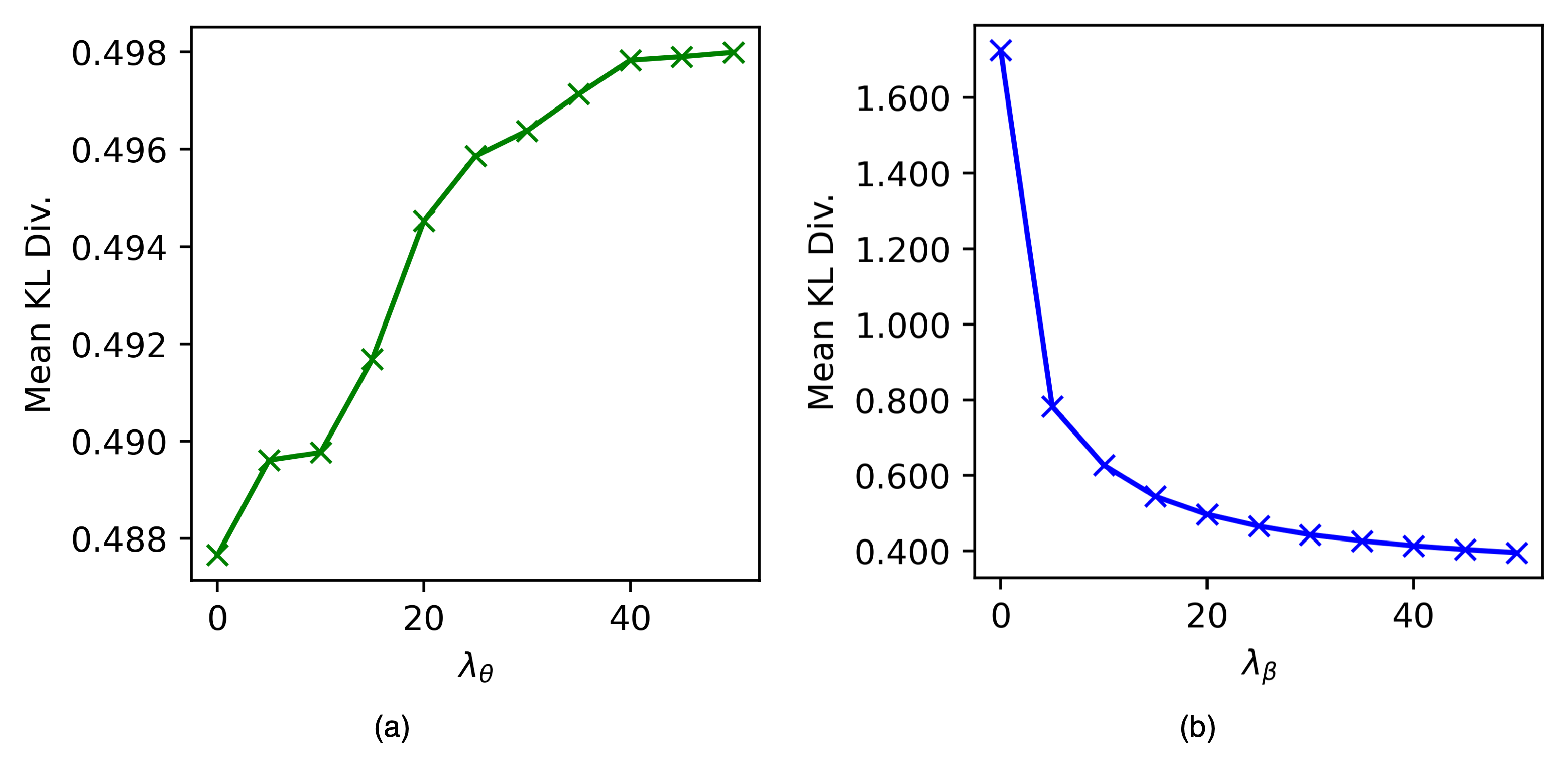}
\caption{Effect of (a) Document-level supervision and (b) Topic-level supervision on topic's adaptability to the target corpus}
\vspace{-5mm}
\label{fig:param_effect}
\end{figure}

%% file: EMNLP 2022/sections/conclusion.tex
\vspace{-1mm}
\section{Conclusion}
\label{sec:conclusion}
\vspace{-1mm}

In this paper, we propose a new problem called coordinated topic modeling that uses a set of well-defined topics to describe a target corpus. It takes a reference representation of topics with their surface names and calibrates those representing the target corpus. The proposed problem describes corpora in a more interpretable and comparable way. To solve the problem, we design an embedding-based coordinated topic model that leverages topic-level and document-level supervision with a self-training mechanism. A set of empirical evaluations demonstrate the superiority of our approach over other strong baselines in multiple datasets.

%% file: EMNLP 2022/sections/acknowledgement.tex
\section*{Acknowledgements}
We thank the reviewers for their constructive feedback.
This material is based upon work supported by the National Science Foundation IIS 16-19302 and IIS 16-33755, Zhejiang University ZJU Research 083650, IBM-Illinois Center for Cognitive Computing Systems Research (C3SR)-- a research collaboration as part of the IBM Cognitive Horizon Network, grants from eBay and Microsoft Azure, UIUC OVCR CCIL Planning Grant 434S34, UIUC CSBS Small Grant 434C8U, and UIUC New Frontiers Initiative. Any opinions, findings, and conclusions or recommendations expressed in this publication are those of the author(s) and do not necessarily reflect the views of the funding agencies.

%% file: EMNLP 2022/sections/limitations.tex
\section*{Limitations}
\label{sec:limitations}

The proposed model can be applied only to those domains where we can find a reference representation of intended topics. However, we aim for the topic's interpretability and comparability with other corpora. Thus, in this work, we are mainly interested in defining a corpus with well-defined topics as input because they are easily understandable to represent a corpus. Moreover, it is very reasonable to think that we can usually obtain reference representation for well-defined topics. 

In this paper, the model assumes that the reference and target corpus are from a common domain. However, it is worth exploring how our model works in cross-domain problems. On the other hand, exploring the proposed model in cross-domain scenarios is also an excellent future direction. We can apply our model in cross-language applications if word embeddings in the vocabulary are available for multiple corpora from different languages.

%% file: EMNLP 2022/sections/appendix.tex
\clearpage
\appendix
\label{sec:appendix}
\input{EMNLP 2022/tables/context_def_full}

\input{EMNLP 2022/sections/related_work}

\section{Additional Text Classification Baselines}
\label{sec:semi-baseline}
\begin{itemize}[nolistsep,leftmargin=*]
    \item SupBERTCLs: It trains a BERT-based supervised text classifier \footnote{\url{https://huggingface.co/docs/transformers/tasks/sequence_classification}} on labeled reference corpus and uses that to predict labels for documents in target corpus.
    \item Zero-ShotCLs: It predicts the topic of each document out of given surface names $\mathcal{C}$ using a Zero-Shot classification \cite{yin2019benchmarking} method based on a pre-trained textual entailment model\footnote{\url{https://huggingface.co/joeddav/bart-large-mnli-yahoo-answers}}
    \item WeSTCLs: This is a weakly-supervised neural text classification method \cite{meng2018weakly} that can classify a text document based on given class names or seed words. We use the two variants: (1) WeSTCLs (Names) uses the topic names $\mathcal{C}$; (2) WeSTCLs (Seeds) uses seed words.
    \item LOTCls: This is a language model-based text classification method \cite{meng2020text} that uses label names as supervision. As supervision, we provide the given topic surface names $\mathcal{C}$.
\end{itemize}

\section{Additional Result on Corpus Comparison}
\label{sec:additional-corpus_comp}

Table \ref{tab:context_words_full} shows qualitative results of corpus comparison. It shows the top 20 words for each corpus generated by our ECTM model, each representing a context for the topic of politics.

%% file: EMNLP 2022/tables/context_def_full.tex
\begin{table*}[!ht]
\resizebox{1.0\linewidth}{!}{%
\begin{tabular}{@{}l|l@{}}
\toprule
Dec. 2000 & \begin{tabular}[c]{@{}l@{}}court, election, vote, bush, president, florida, law, judge, justice, government, \\ supreme, party, clinton, official, case, presidential, decision, ballot, republican, democratic\end{tabular}               \\ \midrule
Sep. 2001 & \begin{tabular}[c]{@{}l@{}}united, attack, war, president, government, official, american, bush, terrorist, military, \\ country, terrorism, national, afghanistan, vote, election, administration, leader, political, muslim\end{tabular} \\ \midrule
USA       & \begin{tabular}[c]{@{}l@{}}republican, vote, campaign, election, president, party, governor, senator, candidate, political, \\ mayor, democratic, clinton, bush, voter, democrat, democrats, giuliani, senate, race\end{tabular}           \\\midrule
Britain   & \begin{tabular}[c]{@{}l@{}}minister, president, ireland, party, government, political, official, prime, irish, \\ northern, clinton, leader, peace, blair, union, thatcher, republican, war, election, sinn\end{tabular}                   \\ \bottomrule

\end{tabular}}
\caption{Top 20 words in politics topic on different contexts}
\label{tab:context_words_full}

\end{table*}

%% file: EMNLP 2022/sections/related_work.tex
\section{Related Work}
\label{sec:related_work}

It is of great interest to automatically mine a set of meaningful and coherent topics for effectively and efficiently understanding and navigating a large text corpus. Topic models \cite{jordan1999introduction,blei2003latent} are such statistical tools that can discover latent semantic themes from a text collection. The main idea is to represent each document in a corpus as a mixture of hidden topics, each represented by word distribution. While most of the early attempts of topic models \cite{griffiths2003hierarchical,blei2006correlated,li2006pachinko} are probabilistic, with the advent of the deep neural network, neural topic models are also proposed. One such example is the Autoencoded Variational Inference For Topic Model (AVITM) \cite{srivastava2017autoencoding}. Recently, Embedded Topic Model (ETM) \cite{dieng2020topic} blends the strengths of the neural topic model and word embedding.

Despite their effectiveness and efficiency, traditional topic models have several limitations. The standard topic models whether it is probabilistic \cite{blei2003latent} or neural \cite{srivastava2017autoencoding,dieng2020topic} has the inability to incorporate user guidance. Existing topic models are typically learned in a purely unsupervised manner and, thus, tend to discover the most general and major topics ignoring user interests. There have been several modifications in the traditional topic models to incorporate user interests or existing knowledge about documents in the corpus. More specifically, there are generally three forms of supervision in existing supervised topic models. 

The first type of supervision is in the form of labeled corpus where all or a subset of documents are annotated with some predefined topic, or class labels \cite{ramage2009labeled,ramage2011partially,mcauliffe2007supervised,lacoste2008disclda}.  
E.g., Labeled LDA \cite{ramage2009labeled} constrains Latent Dirichlet Allocation (LDA) \cite{blei2003latent} model through one-to-one correspondence between LDA’s latent topics and document labels. While they improve the predictive ability of unsupervised topic models, they require a massive set of annotated documents to be effective. The second type of supervision comes in the form of seed words where for each user-interested topic, a set of represented keywords are provided to guide the topic generation process \cite{jagarlamudi2012incorporating,eshima2020keyword,harandizadeh2021keyword,gallagher2017anchored}. More specifically, the seed guided topic models make the topics converge in the direction of given seed words. Although they make the topics converge in the direction of user-desired seed words, they require the seed words to appear in the corpus vocabulary, which may be impractical in many cases. Finally, there is an approach called category-guided topic mining \cite{meng2020discriminative} (CatE), which considers the topics' surface names as the only supervision for mining user-interested discriminative topics. However, it also assumes that the surface name of each topic appears in the corpus. Moreover, none of the above work does impose requirements of making topics comparable over multiple corpora. Moreover, because of strict discrimination assumptions in CatE, they often suffer from mining too specific words to represent topics that are very hard to interpret.

\section{Implementation Details}
\label{sec:implement}
There are some parameters our model we have to set. 
We use the 300-dimensional pre-trained word embedding from Spacy \cite{spacy2}. The dimension of hidden layers in MLP is set to 300. The learning rate is set to 0.005. We use different epochs based on different datasets. We mainly use 150 for news domain datasets and 100 for other cases.
The regularization parameters $\lambda_\beta$ and $\lambda_\theta$ are set to 20 and 35, respectively, for all datasets. We follow the standard procedure for preprocessing text before applying all the models. For example, we remove stopwords; the words appear more than 0.70 of the whole corpus, and infrequent words appear less than ten times. For baselines, we used their default settings.